\pdfoutput=1

\documentclass[11pt]{article}

\usepackage{acl}

\usepackage{algorithm}

\usepackage{times}
\usepackage{latexsym}
\usepackage{xcolor}
\usepackage{soul}
\usepackage{hyperref}
\usepackage{amsmath}
\usepackage{graphicx}
\usepackage{bbm}
\usepackage{booktabs}
\usepackage{multirow}
\usepackage{algpseudocode}
\usepackage{algorithm}
\usepackage{amssymb}
\usepackage[normalem]{ulem}
\usepackage{enumitem}

\usepackage{arydshln}

\usepackage{amssymb}

\usepackage[T1]{fontenc}

\usepackage[utf8]{inputenc}

\usepackage{microtype}

\usepackage{colortbl}

\usepackage{inconsolata}
\usepackage{xcolor}
\definecolor{mycolor1}{HTML}{0000AA}
\definecolor{mycolor2}{HTML}{BB0000}
\definecolor{mycolor3}{HTML}{002200}
\let\svthefootnote\thefootnote
\newcommand\freefootnote[1]{%
  \let\thefootnote\relax%
  \footnotetext{#1}%
  \let\thefootnote\svthefootnote%
}
\newcommand{\methodname}{\textsc{ODSum} }

\newcommand{\geval}{G-E{\small VAL} }

\newcommand{\titlestr}{\methodname: New Benchmarks for Open Domain \\ Multi-Document Summarization}

\title{\titlestr}

\author{
Yijie Zhou$^{\spadesuit\,}$ 
\quad Kejian Shi$^{\clubsuit\, }$ 
\quad Wencai Zhang$^\spadesuit$ \\
\bf{\quad Yixin Liu$^\clubsuit$
\quad Yilun Zhao$^\clubsuit$
\quad Arman Cohan$^{\clubsuit\heartsuit}$} \vspace{8pt}\\
$^\spadesuit$Zhejiang University, $^\clubsuit$Yale University, $^\heartsuit$Allen Institute for AI 
\\
\texttt{\small{e.j.zhou@zju.edu.cn, kejian.shi@yale.edu} }}

\newcommand{\namedref}[2]{\hyperref[#2]{#1~\ref*{#2}}}

\begin{document}
\maketitle
\begin{abstract}

Open-domain Multi-Document Summarization (ODMDS) is a critical tool for condensing vast arrays of documents into coherent, concise summaries.  With a more inter-related document set, there does not necessarily exist a correct answer for the retrieval, making it hard to measure the retrieving performance.
We propose a rule-based method to process query-based document
summarization datasets into ODMDS datasets.  Based on this method, we introduce a novel dataset, ODSum, a sophisticated case with its document index interdependent and often interrelated. We tackle ODMDS with the \textit{retrieve-then-summarize} method, and the performance of a list of retrievers and summarizers is investigated. 
Through extensive experiments, we identify variances in evaluation metrics and provide insights into their reliability. We also found that LLMs suffer great performance loss from retrieving errors.
We further experimented methods to improve the performance as well as investigate their robustness against imperfect retrieval. We will release our data and code at \url{https://github.com/yale-nlp/ODSum}.

\end{abstract}

\section{Introduction}

Summarization is an established NLP task that generates concise and coherent summaries from provided texts~\cite{DBLP:conf/nips/HermannKGEKSB15, Xu2020CoarsetoFineQF, Zhong2021QMSumAN, deyoung-etal-2021-ms, Giorgi2022TowardsMS}. Given the current information surge, the need to extract crucial points from a large collection of documents has become imperative. This introduces the concept of \textit{Open-domain multi-document summarization} (ODMDS)~\citep{Ji2013OpenDomainMS, Giorgi2022TowardsMS}. ODMDS can be analogized to extracting knowledge from diverse pieces of information across a large number of documents, and connecting and aggregate the information into a clear, coherent, and brief summary. While conventional (multi-document) summarization methods typically work within a specified setting where the source documents are predetermined and limited in scope, ODMDS addresses other broader, real-world challenges. It serves as a crucial tool for efficient knowledge extraction, facilitating a more comprehensive understanding of a topic without the need to review a multitude of documents. ODMDS has close connections with information retrieval~\citep{Zhang2021ScalingUQ} and multi-document summarization~\cite{Wallace2020GeneratingN, DeYoung2021MS2MS, Zhong2021QMSumAN}, bridging the gap between extracting relevant documents from large corpora and generating concise summaries from multiple related texts.

\begin{figure}[t!]
    \centering
         \includegraphics[width=0.98\linewidth]{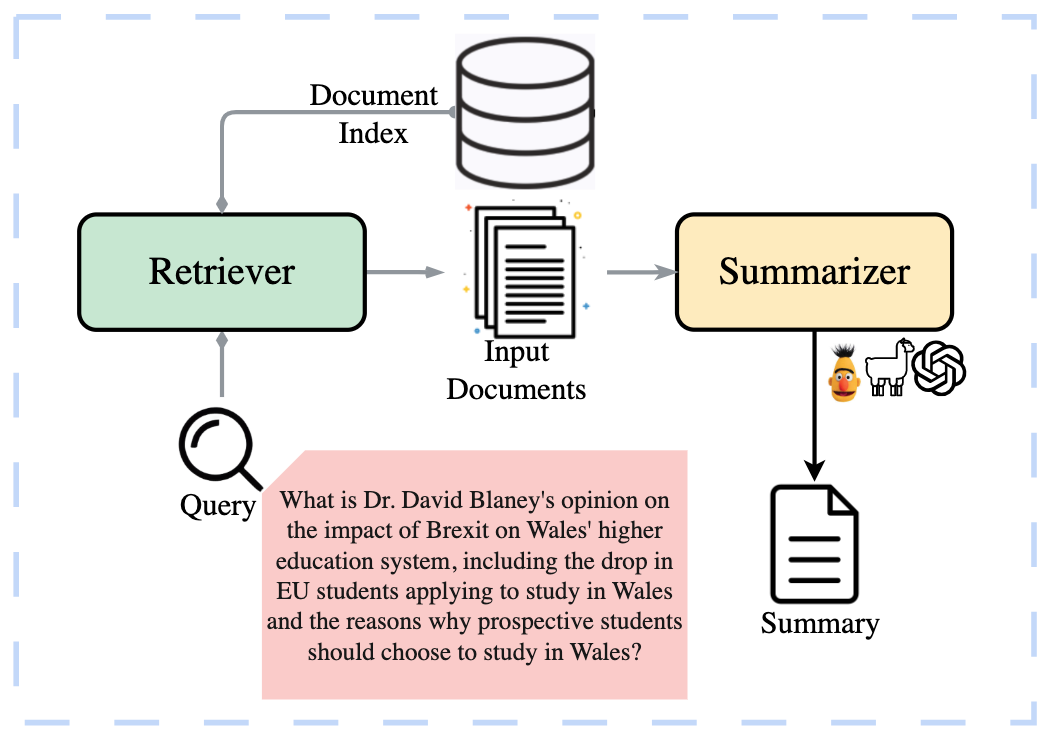}
 \caption{Overview of \emph{retrieve-then-summarize} pipeline for open-domain multi-document summarization task. }
 \label{fig:main_pipeline}
\end{figure}

ODMDS differs from prior work in three main ways: 1) unlike open-domain QA \cite{yang-etal-2015-wikiqa, Lewis2020RetrievalAugmentedGF}, open-domain MDS requires long-form generations that address a given query, 2) unlike Query-focused MDS \cite{Xu2020CoarsetoFineQF, Pasunuru2021DataAF} where the input documents are given and relevant to the query, ODMDS tackles a more challenging scenario where documents must be retrieved from a large-scale collection of mostly irrelevant documents, 3) compared to previous MDS attempts~\cite{Wallace2020GeneratingN, DeYoung2021MS2MS}, which involved hundreds of relevant input documents, ODMDS examines a more challenging and general setting with a large-scale document collection.

Another understudied and intriguing aspect of the ODMDS task arises in more challenging scenarios where there are no unambiguously \textit{correct} documents to retrieve. For example, in a topic with overlapping and redundant information, multiple articles may describe the same issue adequately. Each of these articles could potentially provide a satisfactory answer, making it inherently difficult to evaluate the efficacy of the retrieval process. This quality adds a layer of complexity in measuring the \textit{goodness} of retrieval performance, and that poses challenge to the commonly-applied \textit{retrieve-then-summarize} approach~\cite{Xu2020CoarsetoFineQF, zhang-etal-2022-summn}. However, the performance of subsequent summarization tasks can provide insight into the quality of the retrieved documents. If the summarizer is able to generate coherent and informative summaries, it can serve as an indirect indicator of retrieval performance.

There have been no datasets specifically constructed for the ODMDS task.
This paper seeks to bridge this gap, particularly emphasizing the limitations observed in recent work on ODMDS (\S  \ref{previous work}). Two primary challenges identified in prior studies \cite{Giorgi2022TowardsMS,tang2021improving-pseudo1} include the use of a pseudo-query method (using the reference summary of certain documents as the query) and the sub-optimal construction of the document database. 
To address this limitation, we introduce a novel dataset (\S \ref{sec:ODSum}) tailored to test the limits of ODMDS. 
Our datasets are constructed to satisfy the main requirements of a realistic ODMDS task: (1) they necessitate the need for retrieving and aggregating information across multiple documents; (2) unlike prior work, the queries in our dataset are realistic and represent a user's information need. Our data collection pipeline starts with established datasets like SQuALITY and QMSum, and includes several stages to create instances for our task (\S\ref{sec:ODSum}). 
Our approach not only ensures the authenticity of the query but also fosters a richer, more diverse document index.

An effective system for ODMDS should be equipped to handle retrieval of relevant information from a large collection of documents and address the variability in document quality and relevance to ensure the accuracy and coherence of the final summary. We experiment with a wide range of baselines on our datasets, including task specific MDS focused models, state-of-the-art LLMs, and open-source LLMs. We find that 
based on extensive experiments, we find that the pipeline method we implemented demonstrates the intricate relationship between retrieval and summarization, especially evident through the G-E{\small VAL} metric. Moreover, among the different summarization strategies we tested, there is a considerable variance in performance, underscoring the necessity for a more robust approach to effectively handle the challenges posed by the variability in document quality and relevance in ODMDS tasks.

We conclude our main contributions as follows:
\begin{itemize} [leftmargin=*]
\itemsep0em 
\item We propose a method (\S \ref{q2OD-MDS}) to construct ODMDS datasets from qMDS datasets, and we build ODSum (\S \ref{statistics}) based on two existing datasets (i.e., SQuALITY and QMSum).
\item We conduct extensive experiments on both the retrieval (\S \ref{retieve}) and summarization (\S \ref{sum}) part in the ``retrieve-then-summarize” pipeline, motivating future research on the ODMDS task.
\item We observe that there are dramatic variance between evaluation metrics. The summarization performance is in line with retrieval performance, and that distinction is better observed by certain metrics, such as G-E{\small VAL} (\S \ref{geval}). 
\item We conduct experiments on different summarization strategies (e.g., truncation, map-reduce and refine), and evaluate the performance and robustness of each (\S \ref{sec:ablation_studies}).
\end{itemize}

\section{Related Work}
\label{sec:related}

\subsection{From MDS to query-based MDS}
Multi-Document Summarization (MDS) amplifies the utility of summarization by synthesizing information from diverse sources. Numerous advancements in MDS have been witnessed over the years, addressing challenges related to redundancy elimination, information hierarchy, and graph-driven abstraction, among others~\citep{Yasunaga2017GraphbasedNM, Liu2019HierarchicalTF, Li2020LeveragingGT}. These methods have found extensive application, ranging from news article aggregation~\citep{fabbri-etal-2019-multi} to intricate scientific literature synthesis~\citep{Lu2020MultiXScienceAL} and beyond.

Query-based Multi-Document Summarization (qMDS) is a specialized form of MDS~\cite{Kulkarni2020AQuaMuSeAG, pasunuru2021data, Pasunuru2021DataAF, Zhong2021QMSumAN, zhao2023qtsumm}. It aims at generating a short summary from a set of documents that answers a query, and it demands the ability of discerning information in the text based on specific questions. This makes qMDS inherently more targeted and refined compared to MDS. Large-scale high-quality datasets for qMDS are scarce. AQuaMuSe~\citep{Kulkarni2020AQuaMuSeAG} and QMDSIR~\citep{pasunuru2021data} aimed to address this by automatically retrieving documents with given queries through mining large corpora or web search engine. However, these datasets only include retrieved documents and lacks \textit{gold} documents or references. Hence it is not possible to measure the retrieval performance or study the performance loss due to retrieval errors. QMDSCNN~\citep{Pasunuru2021DataAF} converts existing summarization dataset using the title as a pseudo-query and fetches four presumably related documents. This approach also falls short as it doesn't guarantee the relevance of the retrieved documents. Furthermore, neither of these datasets could be generalized to the setting of ODMDS as they lack a benchmark for comparing ground truth documents with retrieved ones.

\subsection{Previous Attempts on Open-Domain MDS}
\label{previous work}
\citet{Giorgi2022TowardsMS} proposed a two-step process for the ODMDS task: first retrieving relevant documents and then summarizing them. However, two limitations in terms of \emph{pseudo-query} and \emph{dataset construction} are evident in existing approaches.
First, \citet{Giorgi2022TowardsMS} constructed the document index from multiple MDS datasets, which lacked an inherent pairing with a query. To address that challenge, they adopted a pseudo-query method, using the summary itself as a stand-in for a query. However, since the query is an essential feature of this problem set, the pseudo-query approach has inherent limitations that affect both parts of the pipeline. Specifically, in the retriever part, the title may not clearly indicate the focus or scope of the document, and the queries do not actively participate in the summarization process at all, leading to less targeted summaries.
Moreover, \citet{Giorgi2022TowardsMS} constructed the document index by mixing the four data sources together.
A closer examination of which reveals that it provides neither a good coverage nor includes a rich collection of documents.
This limitation makes the retrieval process particularly vulnerable to errors. The database contains insufficient information on each topic to enable robust retrieval. Moreover, the consequence of retrieving an incorrect document is significant due to the sparse nature of the database. For example, \citet{Giorgi2022TowardsMS} attempted to retrieve \textit{10} documents as the max setting in Multi-News, however, the mean number of pertinent documents are only \textit{2.7}.
Given aforementioned two limitations, there is a compelling need for new methodologies that can handle the complexities of ODMDS more effectively.

\subsection{LLMs for Retrieving and Summarizing}

Large language models (LLMs) have exhibited impressive performance in zero-/few-shot tasks across various domains~\cite{Brown2020LanguageMA, Zhang2022OPTOP, Chowdhery2022PaLMSL}.
They have proven to be outperforming strong \emph{retrieval} baselines such as BM25~\citep{bm25} and self-supervised dense retrieval methods~\citep{Brown2020LanguageMA}, and are particularly effective when applied to tasks that lack sufficient supervised data~\citep{Schick2020ItsNJ, Winata2021LanguageMA, Bonifacio2022InParsUD}. 
In terms of \emph{summarization} tasks, recent studies~\citep{goyal2022news, Zhang2023BenchmarkingLL, Yang2023ExploringTL, zhao2023qtsumm} demonstrate that LLMs, when prompted with a task description, produced summaries that were preferred by humans and avoided common dataset-specific issues such as poor factuality.
Furthermore, LLMs also exhibit strong performance in the automated evaluation of summarization~\citep{Shen2023AreLL, Mao2023GPTEvalAS, Liu2023OnLT}.

\section{ODSum Dataset}
\label{sec:ODSum}

%

\subsection{ODMDS Task Setting}
The task of Open-Domain Multi-Document Summarization (ODMDS) can be formally defined as follows: provided a user query $Q$ and a large set of documents $D$, the system is designed to generate a summary $S$. This process can be mathematically depicted as a function $\Phi$ that maps a tuple 
$(D,Q)$, which consists of the set of documents and the query, to a summary $S$. 
The objective of ODMDS is to optimize this function such that the produced summary is both relevant and concise, effectively responding to the query.

\begin{figure*}[!t]
    \begin{minipage}[b]{0.48\linewidth}
            \centering
            \includegraphics[width = \linewidth]{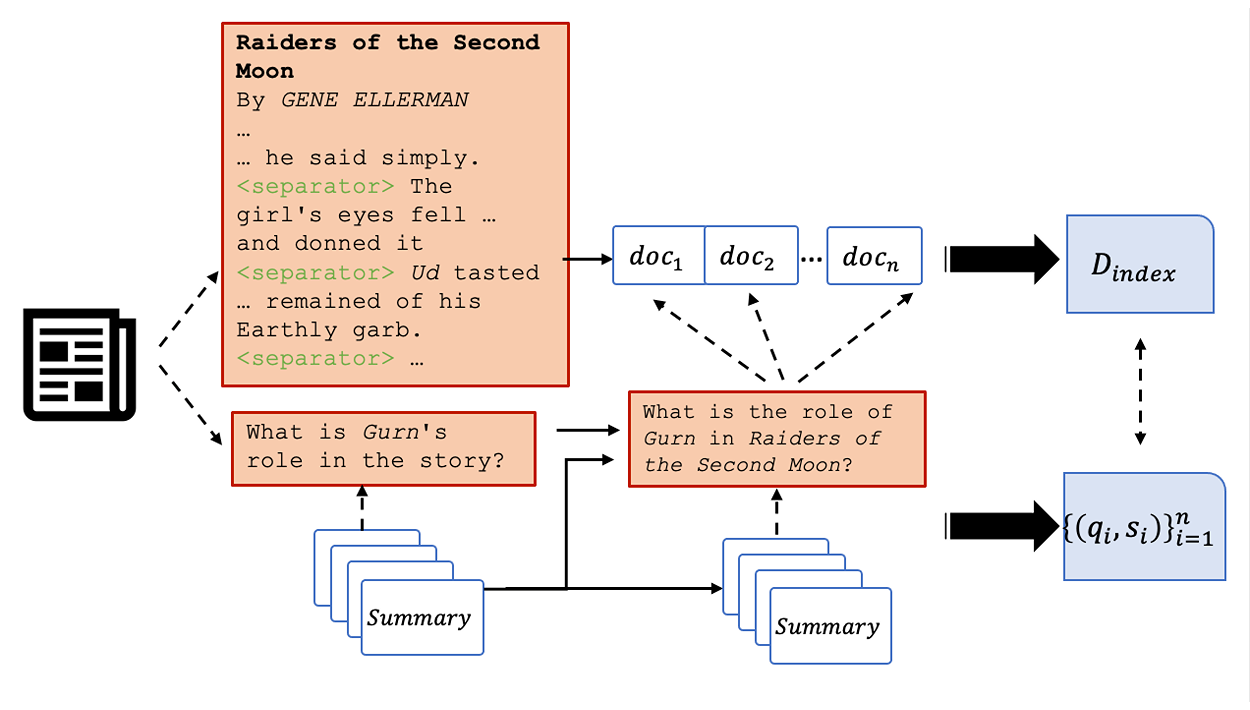}
            \caption*{(a) ODSum-story}
    \end{minipage}\quad
    \begin{minipage}[b]{0.48\linewidth}
            \centering
            \includegraphics[width = \linewidth]{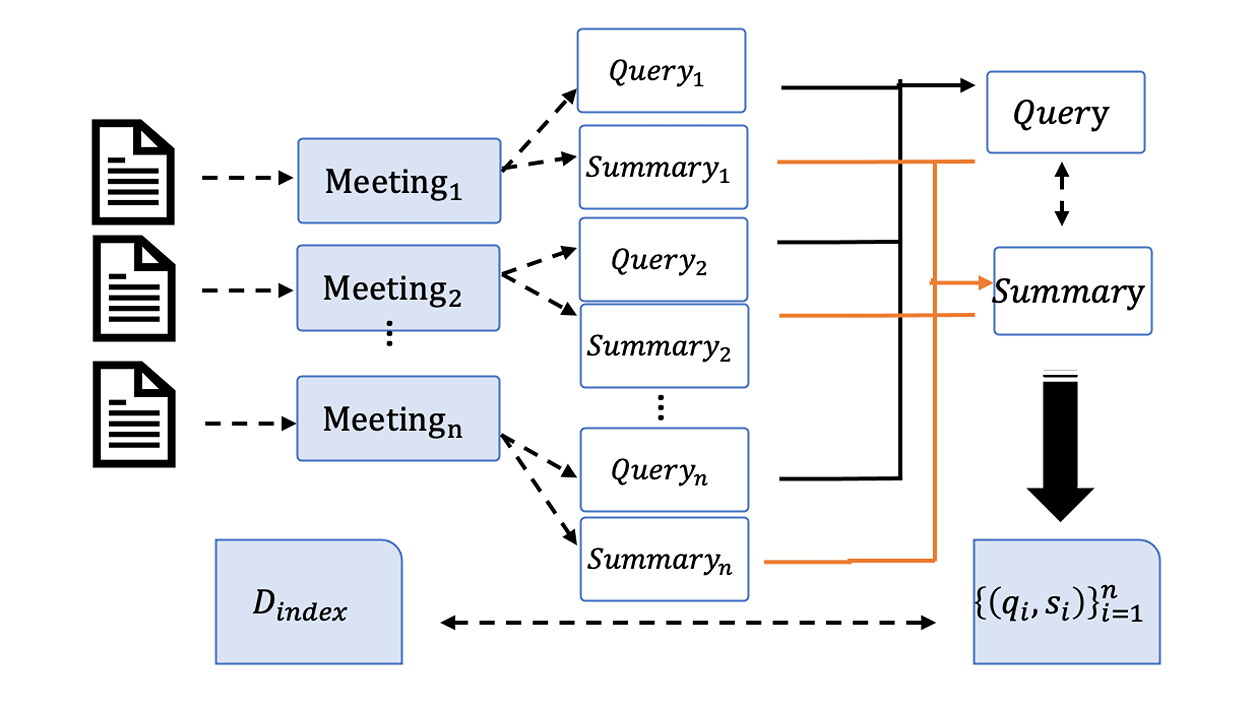}
            \caption*{(b) ODSum-meeting}
    \end{minipage}
    \centering
    \caption{An overview of ODSum dataset construction pipeline.}
    \label{fig:data_story}
    \label{fig:data_meeting}
\end{figure*}

\subsection{Data Source}
We introduce the primary source of data that we used to annotate query-summary pairs: \vspace{3pt}

\noindent\textbf{SQuALITY}~\cite{wang2022squality} is designed for question-based abstractive summarization for short stories. The summaries in the SQuALITY dataset were created by highly-qualified contractors who were hired to read stories and write original summaries from scratch. Each question was answered by four different annotators, who then reviewed each other’s work to ensure the data's high quality. The SQuALITY dataset consists of 127 stories, each of which is associated with 635 questions. These questions were answered in the form of summaries, resulting in a total of 2540 summaries. \vspace{3pt}

\noindent\textbf{QMSum} \citep{Zhong2021QMSumAN} is designed for query-based abstractive summarization of multi-domain meeting transcripts. The summaries are derived from transcribed meetings, focusing on the extraction and summarization of relevant segments in response to specific queries. The dataset comprises 1,808 query-summary pairs that span across 232 meetings from multiple domains. These pairs were meticulously created, ensuring a wide variety of queries and corresponding summaries. 

\subsection{Dataset Construction}
\label{q2OD-MDS}
The open-domain multi-document summarization (ODMDS) task builds upon the query-focused multi-document summarization (qMDS) setting. qMDS involves generating summaries from multiple documents that are relevant to a specific query. However, the scaricity of qMDS datasets remains fundamental challenge. Previous research~\citep{Giorgi2022TowardsMS} attempted to navigate this issue by using document titles as pseudo queries. Nevertheless, queries require targeted, concise summaries that emphasize specific information, beyond what document titles can provide.
To address this shortage, we develop a method to construct high-quality qMDS datasets as a critical \emph{intermediate} step in building ODMDS corpora.
Specifically, we convert existing query-focused \emph{single-document} summarization resources, including SQuALITY \cite{wang2022squality} and QMSum \citep{Zhong2021QMSumAN}, into \emph{multi-document} datasets. 
As demonstrated in Appendix \S \ref{sec:q2ODMDS}, we modify queries by adding contextual details as necessary, ensuring they relate to document collections rather than individual documents. 
In the following subsections, we discuss in detail how we convert SQuALITY and QMSum to the ODMDS setting.



\subsubsection{ODSum-story}
\paragraph{Source Document Split}
Stories are by themselves a chronicle of events. An interesting aspect about SQuALITY is that lots of its stories involve recurring characters, for example, the character \textit{Retief} is a main character in a series of stories that unfold at different times. And since each chapter within a story can be viewed as an independent chronicle occurring at different times, they themselves can be a smaller document capable of telling a story of its own. We utilized the natural delimiters of the stories in its html-scrapped version, specifically web delimiters like \texttt{``<hr class="***"/>''}, to segment the stories into their respective chapters, thereby providing a divided document. The questions in SQuALITY often pertain to multiple chapters of a story, thereby inherently positions our dataset as a qMDS setting.

\paragraph{Contextualize Queries}
The process also involved addressing the inherent ambiguity in some of the queries. When extracted from their original context, these queries were naturally vague and required additional contextualization. To facilitate this, we asked LLMs to rephrase the queries, enhancing their specificity and relevance. We experimented with two different settings to achieve this. One provided with the original query and the title of the story, and the other, the query, the title, and one of the summary to the query. The result of which is listed in Table \ref{tab:query-mod} in Appendix \S\ref{sec:q2ODMDS}. We then selected the best performing queries (modified based both on the title and the answer) to form our new query set. For instance, a generic query like "\textit{What is the plot of the story?}" was transformed into a more context-specific query such as "\textit{What is the story of CULTURAL EXCHANGE and how does it relate to the harvest of Bacchus vines in Lovenbroy and the arrival of exchange students from Bogan?}". This rephrasing provided a more detailed description of the user's intent, thus improving the utility and effectiveness of the queries.

\begin{table}[t!]
\centering 
\setlength\tabcolsep{3pt}
\resizebox{\linewidth}{!}{%
\begin{tabular}{lrr}
\toprule
\textbf{Dataset}  & \textbf{ODSum-story} & \textbf{ODSum-meeting}  \\
\midrule
Documents  &  1,190 & 232        \\
Avg Document Token Length & 808.54 & 7176.21 \\

\noalign{\vskip 0.5ex}\cdashline{1-3}\noalign{\vskip 0.5ex}

Queries & 635 & 436 \\
Avg Query Token Length & 10.79 & 32.89 \\

\noalign{\vskip 0.5ex}\cdashline{1-3}\noalign{\vskip 0.5ex}

Reference Summary per Query  &   4 & 1     \\
Avg Summary Token Length & 273.80 & 185.17 \\
\midrule 
Training Set Size  & 250 & 262 \\
Development Set Size  & 260 & 131\\
Test Set Size  & 125 & 43\\
\toprule
\end{tabular}
}
\caption{Statistics of the constructed ODSum-story and ODSum-meeting dataset. The token length was acquired by \textit{cl100k\_base}\protect\footnotemark.}
\label{tab:dataset-survey}
\end{table}
\footnotetext{\url{https://platform.openai.com/docs/guides/embeddings}}

\subsubsection{ODSum-meeting}
In dealing with the QMSum dataset, we adopted a unique approach that takes advantage of the continuous and topic-focused nature of the meetings that make up the dataset. Each source document in QMSum is a transcript of a meeting, with the meetings often revolving around the same topics as the participants strive to make progress in their discussions. For instance, the 137 meetings in the Product setting where the subject of discussion is the design of a new remote control, the topic of \textit{buttons} featured in 131 of them, \textit{rubber buttons} in 24, and \textit{colored buttons} in 3. If we are interested in what was discussed about \textit{colored buttons}, ideally we would have a query specific to \textit{colored buttons} and a summary that encapsulates the points discussed in the three relevant meetings.

\paragraph{Cluster and Merge Query-Summary Pairs}
To achieve this, we preserved the original meeting transcripts but extracted the queries to form clusters based on their similarity. These similar queries were then merged to form a single, comprehensive query. We applied the same methodology to the summaries, merging them into a single summary that provides a condensed overview of several meetings. The result is a query-summary pair that is naturally aligned with multiple meetings, providing a rich resource for open-domain multi-document summarization tasks. 
A mathematically formalized process can be found in appendix \S \ref{math-def-of-qmsum}.

\subsection{Dataset Statistics}
\label{statistics}
The statistics for each dataset are summarized in Table \ref{tab:dataset-survey}.
Note here that ODSum-story is available with four references per query, and the model's output is compared against multiple ground truths using multi-rouge. In contrast, ODSum-meeting is a more challenging dataset with a significantly higher average token count of 7176.21 per document. This average token length in ODSum-meeting already exceeds most of the input token lengths of common summarization models, necessitating sophisticated techniques for text truncation, segmentation, or hierarchical processing.

\section{Retrieve-then-Summarize Pipeline}
We implement and evaluate the ``retrieve-then-summarize” pipeline for the ODSum tasks.

\subsection{Retrieval Models}
\label{retieve}
Following \citet{Giorgi2022TowardsMS}, we extend the retrieval models to include not only \textit{sparse} and \textit{dense} retrievers but also incorporate \textit{LLM-embedding-based} retrieval into our framework. 

\paragraph{Sparse Retrievers} Sparse retrievers calculates the relevance of a document to a query based on counting the overlapping terms, which are then weighted by their respective frequencies. In this paper we used BM25 \cite{bm25} as a representative in this family.

\paragraph{Dense Retrievers} Dense retrievers embed both documents and queries into a shared embedding space, often times neural language models, and determine their relevance based on the similartiy of the embeddings. This provides a more nuanced and context-aware approach in measuring the relatedness of text strings. In this paper we adopted two different dense retrievers, Contriever \citep{Izacard2021UnsupervisedDI} and GPT-embedding\footnote{\url{https://platform.openai.com/docs/guides/embeddings/limitations-risks}} by openAI. Due to the model having a maximum input token limit of 8191, we explored two alternative approaches when this limit was reached: truncation of the document and taking the weighted average of the subsection of the original document.

\subsection{Summarization Models}
\label{sum}
\textbf{BART}
\citep{lewis-etal-2020-bart} is a sequence-to-sequence Transformer model pre-trained using a sentence permutation objective and a text infilling objective. We use a checkpoint\footnote{\url{https://huggingface.co/facebook/bart-large-cnn}} of \texttt{BART-Large} finetuned on the CNN/DailyMail dataset \citep{hermann-cnndm-2015}. The model has a maximum context length of 1024 tokens.  We finetune the model on ODSum using the AdamW optimizer \cite{loshchilov2019decoupled} for a maximum of 6 epochs with early stopping. Following previous work \citep{wang2022squality}, we format input examples by concatenating a query $q$ to the beginning and the end of the document, separated by an added \texttt{[SEP]} special token. For ODSum-story, every (document, query) input corresponds with four training instances, one for each reference. Due to its limited context length, we consider the model as a simple baseline. \vspace{3pt}

\noindent \textbf{PRIMERA}
 \citep{Xiao2021PRIMERAPM} is a model developed with the explicit aim of handling multi-document summarization tasks. 
 It employs efficient encoder-decoder transformers that simplify the processing of concatenated input documents, offering a streamlined approach to MDS. In this task, we finetune the PRIMERA model on each setting of our dataset. Because the max input token length is 4K, we truncate each retrieved document to fit in the context limitation.

\noindent \textbf{GPT} \cite{Brown2020LanguageMA} is a large language model introduced by OpenAI. It has demonstrated significant potential and value in the classical NLP task of text summarization. In this task, we utilized both \texttt{gpt-3.5-16k-turbo-0613} and \texttt{gpt-4-0613}\footnote{\url{https://platform.openai.com/docs/models/overview}} to perform the summarization task. This required the careful design of appropriate prompts to correctly instruct the GPT models to carry out the task. Throughout this process, after designing and testing a variety of prompts, we found that GPT models generally prefer concise, shorter prompts and usually perform better in such scenarios. We also observed that when a query is placed at the end of an article, the resulting summary tends to be of higher quality. This is perhaps because GPT tends to retain the most recent text and may forget preceding context. Our designed prompts are listed in Appendix \S \ref{sec:template}. 
Once the prompts were designed, we truncated stories/meetings to suitable lengths to fit within the maximum input tokens of GPT-3.5 (16k tokens) and GPT-4 (8k tokens). \vspace{3pt}

\noindent \textbf{Llama-2}
\citep{Touvron2023Llama2O} is a collection of auto-regressive text models that has exhibited versatility ranging from logical reasoning to text generation. We use the 70 billion-parameter \texttt{Llama-2-70b-Chat} checkpoint, which was optimized for dialog use cases. At inference time, we load the model weights in 4-bit using NF4 quantization \cite{dettmers2023qlora}.

\begin{table*}[t!]
\footnotesize
\begin{center}
\begin{tabular}{lllrrrr}  %
 \toprule
 Dataset & Top-k & Method & P@K & R@K & NDCG & MAP \\
 \midrule
 \multirow{9}{*}{ ODSum-story } 
 & min(3) & BM25 & 45.88  & 24.44 & 45.29  & 14.76 \\
  &  & Contriever & 52.65 & 27.48 &   53.00  & 17.08 \\ 
  &  & GPT-embedding & \textbf{67.14}  & 35.70 &  68.33  & 23.29 \\
\noalign{\vskip 0.7ex}\cline{2-7}\noalign{\vskip 0.7ex}
  & mean(8) & BM25 & 30.00 & 28.50 &   32.49  & 18.92 \\
  &  & Contriever & 38.13 & 35.39 &   41.51  & 24.51\\
  &  & GPT-embedding & 37.19 & \textbf{41.62} & \textbf{48.43}   &  \textbf{30.13} \\
\noalign{\vskip 0.7ex}\cline{2-7}\noalign{\vskip 0.7ex}
  & max(10) & BM25 & 26.16 & 27.56 &  30.53  & 18.35 \\
  &  & Contriever & 33.84 & 34.88 & 39.23  & 24.31 \\
  &  & GPT-embedding & 38.61 & 40.12 &  44.98  & 29.15 \\
 \midrule
 \multirow{9}{*}{ ODSum-meeting } 
& min(1) & BM25 & \textbf{29.82}  & \textbf{20.16} & 29.82  & 16.94\\
  &  & Contriever & 15.83 & 11.08 & 15.83  & 9.53\\
  &  & GPT-embedding & 25.00  & 16.51 & 25.00  & 13.71\\
\noalign{\vskip 0.7ex}\cline{2-7}\noalign{\vskip 0.7ex}
  & mean(3) & BM25 & 18.12 & 20.03 & 28.14  & 22.71\\
  &  & Contriever & 11.47& 12.82 &17.40  &13.37\\
  &  & GPT-embedding & 17.28& 18.62 &25.26&19.74\\
\noalign{\vskip 0.7ex}\cline{2-7}\noalign{\vskip 0.7ex}
  & max(6) & BM25 & 13.04 & 17.98 &\textbf{31.57} & \textbf{25.82}\\
  &  & Contriever & 8.45& 11.68 & 19.78& 15.09\\
  &  & GPT-embedding & 12.23 & 16.68 & 28.09 & 22.31\\
\bottomrule
\end{tabular}
\end{center}
\caption{Retrieval performance.}
\label{tab:retrieve}
\end{table*}

\section{ODSum Evaluation}
\subsection{Retrieval Evaluation}
We report the precision and recall at \textit{k} (P@K and R@K); P@K is the fraction of the top-k retrieved documents considered relevant, and R@K is the fraction of known relevant documents appearing in the top-k retrieved results. We also report NDCG (Normalized Discounted Cumulative Gain) for understanding how well the most relevant documents are ranked at the top, and MAP (Mean Average Precision) to give a single-figure measure of quality across recall levels. Each metric offers a different perspective on the system's performance, helping us understand its strengths and weaknesses.
\subsection{Summarization Evaluation}
We evaluated the quality of summaries generated by different models using following three metrics: \vspace{3pt}

\noindent \textbf{ROUGE} \cite{lin-2004-rouge} measures the word overlap between the candidate and reference summaries. We reported F1 score for ROUGE-2. \vspace{3pt}

\noindent \textbf{BERTScore} \cite{BERTScore} calculates the similarity between the reference and generated summary using contextual word embeddings. We chose F1 score as evaluation metric. \vspace{3pt}

\noindent \textbf{G-E{\small VAL}} \cite{Liu2023GEvalNE} is a framework that employs LLMs with a chain-of-thoughts (CoT) approach and a form-filling paradigm to assess the quality of NLG outputs. Given the original text, questions, and generated answers, the GPT model is instructed to provide scores across various dimensions (e.g., consistency, coherence, relevance, and fluency) based on the given prompts.

\subsubsection{\geval Implementation Details}
We employed \texttt{gpt-3.5-turbo-16k-0613} as the backbone of \geval to compare the consistency and relevance between the predicted summary and the reference summary. The rationale behind this approach is as follows:
\begin{itemize} [leftmargin=*]
\itemsep0em 
    \item Popular current evaluation metrics, such as ROUGE~\cite{lin-2004-rouge} and BERTScore~\cite{BERTScore}, assess the similarity between the predicted summary and the reference summary without taking the source input into account. Consequently, G-E{\small VAL} should align with this standard.
    \item The constraints imposed by GPT's max input tokens make it impossible to input the original text, question, and generated summary simultaneously. Additionally, inputting only the reference summary can be more cost-effective.
    \item When a prediction aligns with a reference and guarantees both the accuracy and completeness of the information, it is deemed a good prediction. Therefore, we compared the consistency and relevance of the two summaries, scoring them based on these criteria.
\end{itemize}

Based on the aforementioned considerations, we carefully designed the \geval prompts for the ODSumm task, which an be found in appendix \S \ref{sec:template}. After scoring each example, we computed the average, using this mean value as an indicator of the quality of the model-generated summary.
\section{Experiments and Analysis}
\subsection{Analysis of Retriever Performance}

In this section, we discuss the retrieval experiments conducted to evaluate the performance of various retrieval methods (Table \ref{tab:retrieve}).

\subsubsection{ODSum-story Retriever Performance}

The performance metrics on the ODSum-story dataset showed that the dense retriever performed better than the sparse retriever, and that among dense retrievers, LLM-embedding-based performed better than contriever-based. LLM model attained the highest scores across all experiments. The performance gains can likely be attributed to the more nuanced, context-aware embedding that the LLM model can generate.

\subsubsection{ODSum-meeting Retriever Performance}

Contrary to the results on ODSum-story setting, sparse retrieval methods showed higher performance on the ODSum-meeting in terms of all the scores. 
We suspect this is due to the complex nature of the structure of the ODSum-meeting dataset, and this evaluation method may not well demonstrate the quality of the retriever.

 As elaborated in \S \ref{q2OD-MDS}, this dataset involves query-summary pairs that span multiple meeting transcripts, often revolving around intricately related topics. Compared to the ODSum-story dataset which has simpler, more isolated texts, the multi-domain meeting transcripts in QMSum are interdependent and often interrelated. 

It is worth noting that QMSum presents documents that relate to each other and not necessarily to the original query alone. 
Because BM-25 is essentially a bag-of-words approach~\citep{bm25} that doesn't consider the semantics or the context in which the words appear, BM-25 may perform better in retrieving more documents that are "originally related" to the query, but it does not necessarily make it a better retriever in this intricate scenario. On the other hand, dense retrievers like Contriever and GPT-embedding capture semantic meanings of the text. 
This hypothesis is supported in next section (\S \ref{sec:summresult}), where the ultimate summarization performance of dense retrievers actually outperform sparse retrievers by a large extent. This suggested that the ultimate summarization score with the same model should be a better indication of which retriever performs better.

\begin{table*}
  \centering
  \setlength\tabcolsep{3pt}
\resizebox{\textwidth}{!}{%

  \begin{tabular}{@{}lcccccccccccccccccccc@{}}
    \toprule
     &  \multicolumn{3}{c}{BART} & \multicolumn{3}{c}{PRIMERA} & \multicolumn{3}{c}{GPT-3.5} & \multicolumn{3}{c}{GPT-4} & \multicolumn{3}{c}{LLAMA2 - 70B} \\
    \cmidrule(lr){2-4} \cmidrule(lr){5-7} \cmidrule(lr){8-10} \cmidrule(lr){11-13} \cmidrule(lr){14-16}
     &    R-2 & BS & \geval & R-2 & BS & \geval & R-2 & BS & \geval & R-2 & BS & \geval & R-2 & BS & \geval \\
    \midrule
    Oracle  & 11.57 & 84.55 & 20.59 & 10.63 & 86.10 & 20.46 & 12.28 & 86.26 & \textbf{39.66} & 10.24 & 85.68 & \cellcolor{yellow!46}{\textbf{46.54}}  & 10.75 & 85.72 & \textbf{35.62} \\
    \midrule
    \textit{GPT-embedding} 
         & 11.28 & 84.63 & 20.71 & 10.12 & 84.56 & 17.56 & 10.18 & 85.19 & 35.24 & 6.98 & 84.46 & \cellcolor{yellow!39}{38.90} & 9.23  & 85.09 & 34.64 \\

    \midrule
    Contriever    & 10.84 & 84.59 & 19.37 & 10.53 & 85.95 & 19.00 & 9.34  & 84.87 & 31.80 & 6.76 & 84.48 & \cellcolor{yellow!36}{36.63} & 8.63  & 84.81 & 32.37 \\
    \midrule
    BM-25 
          & 10.91 & 84.46 & 18.71 & 10.18 & 84.23 & 18.21 & 9.15  & 84.91 & 31.29 & 5.33 & 84.21 & \cellcolor{yellow!34}{34.18} & 8.77  & 84.78 & 31.27 \\
    \bottomrule
  \end{tabular}
  }
    \caption{Summarization performance on ODSum-story. R-2 and BS stands for ROUGE-2 and BERTScore.}
\label{tab:SummarizationS}
\end{table*}

\begin{table*}
  \centering
  \setlength\tabcolsep{3pt}
    \resizebox{\textwidth}{!}{%

  \label{tab:SummarizationQ}
  \begin{tabular}{@{}lccccccccccccccccccc@{}}
    \toprule
     &  \multicolumn{3}{c}{BART} & \multicolumn{3}{c}{PRIMERA} & \multicolumn{3}{c}{GPT-3.5} & \multicolumn{3}{c}{GPT-4} & \multicolumn{3}{c}{LLAMA2 - 70B} \\
    \cmidrule(lr){2-4} \cmidrule(lr){5-7} \cmidrule(lr){8-10} \cmidrule(lr){11-13} \cmidrule(lr){14-16}
     &  R-2 & BS & \geval & R-2 & BS & \geval & R-2 & BS & \geval & R-2 & BS & \geval & R-2 & BS & \geval \\
    \midrule
    Oracle 
         & 11.68 & 86.35 & 25.65 & 10.04 & 85.86 & 23.85 & 11.89 & 85.30 & \textbf{33.68} & 8.73 & 84.50 & \cellcolor{yellow!36}{\textbf{36.39}} & 8.36 & 85.05 & \textbf{32.26} \\
    \midrule
    \textit{text-ada-002} 
          & 10.86 & 86.21 & 23.17 & 10.46 & 85.31 & 23.59 & 9.88 & 85.31 & 31.11 & 7.48 & 84.53 & \cellcolor{yellow!30}{35.32} & 8.03 & 84.82 & 29.96 \\
    \midrule
    Contriever 
          & 11.10 & 86.22 & 23.67 & 10.62 & 84.69 & 22.10 & 8.15 & 84.69 & 28.18 & 8.76 & 84.72 & \cellcolor{yellow!26}{32.60} & 8.17 & 84.90 & 28.28 \\
        \midrule
    BM-25 
           & 10.83 & 86.30 & 24.05 & 9.63 & 85.91 & 22.48 & 6.96 &  83.71 & 24.53 & 6.01 & 84.10 &\cellcolor{yellow!15}{25.57} & 7.32 & 84.48 & 27.25 \\

    \bottomrule
  \end{tabular}
  }
    \caption{Summarization performance on ODSum-meeting. R-2 and BS stands for ROUGE-2 and BERTScore.}
\label{tab:SummarizationQ}
\end{table*}

\subsection{Summarization Experiments and Analysis}
\label{sec:summresult}

We list the results of summarization in Table \ref{tab:SummarizationS} and Table \ref{tab:SummarizationQ}. The rows are ranked from generally the best performance to the worst.

\subsubsection{Discerning Retrieval Errors through Evaluation Metrics}
\label{geval}

We outline key observations to understand how the performance of summarization models is impacted by retrieval errors and how well these variances are captured by different evaluation metrics.

\paragraph{\geval is more reliable for ODSum evaluation}

Models exhibit little variation in both ROUGE and BERTScore scores.
For instance, GPT-4, widely regarded as a superior iteration with enhancements across various dimensions~\cite{goyal2022news, zhang-etal-2022-summn}, posts unexpectedly low scores based on these metrics.
Furthermore, even when comparing the same model using different retrieval methods, there's only a slight fluctuation in ROUGE and BERTScore. Surprisingly, the performance of documents with errors sometimes surpasses that of the oracle setting.
Such findings cast doubt on the reliability of these metrics in capturing the nuances of summarization quality.
In contrast, the \geval metric shows considerable variability across different models and retrieval settings. GPT-4 excels according to \geval, distancing itself from other LLMs and pretrained models. \geval also reveals a significant performance gap between Oracle settings and other retrieval methods. For instance, the Oracle setting for GPT-4 in ODSum-story outperforms the worst retrieval method by 12.36. Such findings demonstrate that \geval provides a more nuanced evaluation landscape for the ODMDS task, especially when retrieval errors are a consideration. Therefore, we analyze the impact of retrieval errors using \geval.

\subsubsection{Impact of Retrieval Errors}
One of the crucial observations from our experiments is that the quality of summarization is directly subject to retrieval errors. For the ODSum-story setting, as exhibited in the Table \ref{tab:SummarizationS}, the Oracle setting, which represents an ideal retrieval scenario, outperforms all other retrieval settings (BM-25, Contriever, and \textit{text-ada-002}). Also, the performance ranking of the summarization models aligns closely with that of the retrieval methods, following the pattern: \textit{text-ada-002} > Contriever > BM-25. This suggests that the performance of the summarization model can significantly vary based on the quality of the retrieved documents, thereby making retrieval a bottleneck for effective summarization.

For the ODSum-meeting setting in Table \ref{tab:SummarizationQ}, a similar trend is to be observed, with the dense retriever outperforming the sparse retriever. This backs our analysis in \S \ref{sec:dataset_Example} that even though for each query BM-25 retrieved more documents that its originally related, it still doesn't make it better retriever, because ODSum-meeting presents a more complex picture with its documents relating with each other and not necessarily to its own, and that simply evaluating the performance with the ground truth documents does not suffice to measure the retrieval performance.

\subsubsection{LLM's Sensitivity to Retrieval Errors}
\label{gpt4sensitivity}
When retrieval errors are introduced, the GPT-4 model experiences a significant decline in its performance scores across both the story and meeting settings, from 46.54 in the Oracle setting to 34.18 with BM-25 retrieval in ODSum-story. On the meeting setting, from 36.39 to 25.57. 
For comparison, GPT-3.5-turbo-16k and LLAMA2-70B display a similar performance in the oracle setting, among the two, LLAMA2-70B seems to exhibit a higher level of robustness, maintaining more of its performance despite the introduction of retrieval errors.

This sensitivity of GPT-4 to retrieval errors underscores the importance of considering the context in which a summarization model will be deployed. While GPT-4 may outperform other models in ideal conditions, its effectiveness may be considerably hampered when those conditions are not met. This observation further highlights the utility of the G-E{\small VAL} metric, which is able to capture such nuances in model performance that are otherwise glossed over by traditional metrics like ROUGE and BERTScore.

\section{What methods can we take to better the performance?}

\label{sec:ablation_studies}

In an effort to enhance both the MDS performance and the robustness of LLM against errors in information retrieval, we conduct ablation studies focusing on the best and worst performing retrieval settings. Specifically, we used the ODSum-meeting dataset and compared the ideal scenario, represented by the oracle setting, with the worst-performing setting (sparse) and the MAX to understand their differences and impact on LLM's performance. Both setting has documents with very long token length, exceeding the maximum input limits of the gpt-3.5-turbo-16k here we choose to experiment on, and methods other than simply truncate the inputs are experimented here.

\subsection{Methods to Improve Robustness}
Several methods are investigated to make the LLM more resilient against retrieval errors, inspired the framework presented in \textit{LangChain}:

\begin{itemize}[leftmargin=*]
\itemsep0em 
\item \textbf{Truncate Methods:} Due to the token limitations of LLMs, two variations of the truncate method are employed to manage the document size.

\textbf{Method 1 - Truncate all:} In this approach, all retrieved documents are concatenated together and then truncated. This often leaves the model with just the initial documents, which are typically the most relevant according to the retriever's ranking. 

\textbf{Method 2 - Truncate one:} Each individual document is truncated first. The truncated parts from each document are then combined together to form a new document that fits within the model's token limit.

\item \textbf{Map-Reduce Method:} This architecture allows for summarizing large collections of documents by distributing work across mappers and reducers. In this approach, the document is divided into smaller chunks that conform to the token limit of the LLM. Each of these chunks is initially summarized using an initial prompt (serving as the map function). Subsequently, these intermediate summaries are combined into a final summary using a second, different prompt (serving as the reduce function). This method offers scalability, however information may be lost during the intermediate step.

\item \textbf{Refine Method:} This technique iteratively updates the generated summary by sequentially processing each retrieved document. For each round, the current document along with the latest interim summary is input into an LLM to generate a revised summary. This method is also scalable, however, the order of the document might be crucial for its performance. Here we try two type of document orders, ranking highest to lowest or vice versa, according to relevancy.

\end{itemize}

\begin{table}[t]

\begin{center}
\scalebox{0.8}{
\begin{tabular}{ lrrrrr } 
 \toprule
 Method & setting & r1 & r2 & rL & g-eval\\
 \midrule
 Truncate\_all & oracle & 40.20 & 11.89 & 20.62 & 33.68 \\
& \textit{with error} & \textit{-8.51}  & \textit{-4.93} & \textit{-3.10} &  \textit{-9.15}\\

\noalign{\vskip 0.5ex}\cdashline{1-6}\noalign{\vskip 0.5ex}

 Truncate\_one & oracle & 40.03 & 11.41 & 20.72 & 35.42 \\
& \textit{with error} & \textit{-6.70}  & \textit{-3.81} & \textit{-2.48} &  \textit{-10.94}\\

\noalign{\vskip 0.5ex}\cdashline{1-6}\noalign{\vskip 0.5ex}

 Map-Reduce & oracle & 35.90 & 10.72 & 18.90  & \textbf{37.70} \\
 & \textit{with error} & \textit{-3.97} & \textit{-3.99} & \textit{-1.72}  & \textit{-12.16}\\

 \noalign{\vskip 0.5ex}\cdashline{1-6}\noalign{\vskip 0.5ex}
 
 Refine\_H2L & oracle & 38.23 & 10.60 & 19.48& 36.36\\
&  \textit{with error} &\textit{-4.87} & \textit{-2.51}& \textit{-2.55}& \textit{-10.23}\\

\noalign{\vskip 0.5ex}\cdashline{1-6}\noalign{\vskip 0.5ex}

 Refine\_L2H & oracle & 42.23 &11.19 &21.45& 34.48 \\
&  \textit{with error} & \textit{-7.37} & \textit{-4.06}& \textit{-4.84}& \textit{-9.26}\\
\bottomrule
\end{tabular}
}
\end{center}
\caption{Different methods to summarize documents with retrieval errors.}
\label{tab:ablation}
\end{table}

\subsection{Discussion}

The results presented in Table \ref{tab:ablation} shed light on various strategies for improving the robustness of LLM against retrieval errors, particularly in the context of the ODSum-meeting dataset.

In the oracle setting, \textit{Truncate\_all} performs marginally better than the \textit{Truncate\_one} in the oracle setting, and \textit{Map-reduce} have the best score among all settings listed in the table. This suggests that maintaining the diversity of information by whether summarizing or truncating all the document might be more beneficial than merely keeping the supposedly more relevant initial documents. 

While \textit{Map-reduce} performs the best in terms of G-E{\small VAL}, it has the lowest ROUGE score. This can be likely attributed to this method has at least a two-level abstraction of word selections and introduced more novel words. This result further backs our observation in \S \ref{sec:ablation_studies} that ROUGE score does not ideally reflect the summary quality.

Both variants of the Refine Method show interesting results. The \textit{Refine\_H2L} (High to Low relevance) performs in G-E{\small VAL}, and \textit{Refine\_L2H} (Low to High relevance) has the highest ROUGE score in the oracle setting. Clearly the order in which documents are processed is critical, the model is likely to use the words appeared in the most recent document, giving \textit{Refine\_L2H} the highest ROUGE score, but \textit{Refine\_H2L} has a better summary.

From a broader perspective, all of the methods proposed to include more information for the model achieved a better score in the oracle setting. This suggests that providing more information is generally beneficial for the model's summarization capabilities when the retrieval process is ideal. However, the downside becomes evident when retrieval errors are introduced. Not only do these advanced methods suffer a significant drop in performance, but surprisingly, the degradation is even more pronounced than what is experienced by the simpler Truncate\_all method itself. Therefore, these findings invite further exploration into methods that not only maximize the inclusion of relevant information but can also contain the negative impacts of retrieval errors. Simple truncation methods may have a sort of "natural resilience" to these errors because they operate under constraints that limit their exposure to irrelevant information. Advanced methods may need to incorporate similar safeguarding strategies to improve their robustness.

\section{Conclusion}

This paper introduces a novel dataset, ODSum, for the ODMDS task. Our comprehensive experimental approach covers both retrieval and summarization stages in the "retrieve-then-summarize" pipeline, highlighting areas for further investigation in the ODMDS task. Through our analysis, we observed significant variations among evaluation metrics. We found that the correlation between summarization and retrieval performance is best highlighted by certain metrics, particularly \geval. Moreover, our research investigates the effectiveness and robustness of various summarization strategies, including truncation, map-reduce, and refine, providing insights into their individual advantages and limitations.


\begin{thebibliography}{46}
\expandafter\ifx\csname natexlab\endcsname\relax\def\natexlab#1{#1}\fi

\bibitem[{Bonifacio et~al.(2022)Bonifacio, Abonizio, Fadaee, and Nogueira}]{Bonifacio2022InParsUD}
Luiz~Henrique Bonifacio, Hugo Abonizio, Marzieh Fadaee, and Rodrigo Nogueira. 2022.
\newblock \href {https://api.semanticscholar.org/CorpusID:250340449} {Inpars: Unsupervised dataset generation for information retrieval}.
\newblock \emph{Proceedings of the 45th International ACM SIGIR Conference on Research and Development in Information Retrieval}.

\bibitem[{Brown et~al.(2020)Brown, Mann, Ryder, Subbiah, Kaplan, Dhariwal, Neelakantan, Shyam, Sastry, Askell, Agarwal, Herbert-Voss, Krueger, Henighan, Child, Ramesh, Ziegler, Wu, Winter, Hesse, Chen, Sigler, Litwin, Gray, Chess, Clark, Berner, McCandlish, Radford, Sutskever, and Amodei}]{Brown2020LanguageMA}
Tom~B. Brown, Benjamin Mann, Nick Ryder, Melanie Subbiah, Jared Kaplan, Prafulla Dhariwal, Arvind Neelakantan, Pranav Shyam, Girish Sastry, Amanda Askell, Sandhini Agarwal, Ariel Herbert-Voss, Gretchen Krueger, T.~J. Henighan, Rewon Child, Aditya Ramesh, Daniel~M. Ziegler, Jeff Wu, Clemens Winter, Christopher Hesse, Mark Chen, Eric Sigler, Mateusz Litwin, Scott Gray, Benjamin Chess, Jack Clark, Christopher Berner, Sam McCandlish, Alec Radford, Ilya Sutskever, and Dario Amodei. 2020.
\newblock \href {https://api.semanticscholar.org/CorpusID:218971783} {Language models are few-shot learners}.
\newblock \emph{ArXiv}, abs/2005.14165.

\bibitem[{Chowdhery et~al.(2022)Chowdhery, Narang, Devlin, Bosma, Mishra, Roberts, Barham, Chung, Sutton, Gehrmann, Schuh, Shi, Tsvyashchenko, Maynez, Rao, Barnes, Tay, Shazeer, Prabhakaran, Reif, Du, Hutchinson, Pope, Bradbury, Austin, Isard, Gur-Ari, Yin, Duke, Levskaya, Ghemawat, Dev, Michalewski, Garc{\'i}a, Misra, Robinson, Fedus, Zhou, Ippolito, Luan, Lim, Zoph, Spiridonov, Sepassi, Dohan, Agrawal, Omernick, Dai, Pillai, Pellat, Lewkowycz, Moreira, Child, Polozov, Lee, Zhou, Wang, Saeta, D{\'i}az, Firat, Catasta, Wei, Meier-Hellstern, Eck, Dean, Petrov, and Fiedel}]{Chowdhery2022PaLMSL}
Aakanksha Chowdhery, Sharan Narang, Jacob Devlin, Maarten Bosma, Gaurav Mishra, Adam Roberts, Paul Barham, Hyung~Won Chung, Charles Sutton, Sebastian Gehrmann, Parker Schuh, Kensen Shi, Sasha Tsvyashchenko, Joshua Maynez, Abhishek Rao, Parker Barnes, Yi~Tay, Noam~M. Shazeer, Vinodkumar Prabhakaran, Emily Reif, Nan Du, Benton~C. Hutchinson, Reiner Pope, James Bradbury, Jacob Austin, Michael Isard, Guy Gur-Ari, Pengcheng Yin, Toju Duke, Anselm Levskaya, Sanjay Ghemawat, Sunipa Dev, Henryk Michalewski, Xavier Garc{\'i}a, Vedant Misra, Kevin Robinson, Liam Fedus, Denny Zhou, Daphne Ippolito, David Luan, Hyeontaek Lim, Barret Zoph, Alexander Spiridonov, Ryan Sepassi, David Dohan, Shivani Agrawal, Mark Omernick, Andrew~M. Dai, Thanumalayan~Sankaranarayana Pillai, Marie Pellat, Aitor Lewkowycz, Erica Moreira, Rewon Child, Oleksandr Polozov, Katherine Lee, Zongwei Zhou, Xuezhi Wang, Brennan Saeta, Mark D{\'i}az, Orhan Firat, Michele Catasta, Jason Wei, Kathleen~S. Meier-Hellstern, Douglas Eck, Jeff Dean, Slav Petrov,
  and Noah Fiedel. 2022.
\newblock \href {https://api.semanticscholar.org/CorpusID:247951931} {Palm: Scaling language modeling with pathways}.
\newblock \emph{ArXiv}, abs/2204.02311.

\bibitem[{Dettmers et~al.(2023)Dettmers, Pagnoni, Holtzman, and Zettlemoyer}]{dettmers2023qlora}
Tim Dettmers, Artidoro Pagnoni, Ari Holtzman, and Luke Zettlemoyer. 2023.
\newblock \href {http://arxiv.org/abs/2305.14314} {Qlora: Efficient finetuning of quantized llms}.

\bibitem[{DeYoung et~al.(2021{\natexlab{a}})DeYoung, Beltagy, van Zuylen, Kuehl, and Wang}]{deyoung-etal-2021-ms}
Jay DeYoung, Iz~Beltagy, Madeleine van Zuylen, Bailey Kuehl, and Lucy~Lu Wang. 2021{\natexlab{a}}.
\newblock \href {https://doi.org/10.18653/v1/2021.emnlp-main.594} {{MS}{\^{}}2: Multi-document summarization of medical studies}.
\newblock In \emph{Proceedings of the 2021 Conference on Empirical Methods in Natural Language Processing}, pages 7494--7513, Online and Punta Cana, Dominican Republic. Association for Computational Linguistics.

\bibitem[{DeYoung et~al.(2021{\natexlab{b}})DeYoung, Beltagy, van Zuylen, Kuehl, and Wang}]{DeYoung2021MS2MS}
Jay DeYoung, Iz~Beltagy, Madeleine van Zuylen, Bailey Kuehl, and Lucy~Lu Wang. 2021{\natexlab{b}}.
\newblock \href {https://api.semanticscholar.org/CorpusID:233231380} {Msˆ2: Multi-document summarization of medical studies}.
\newblock \emph{ArXiv}, abs/2104.06486.

\bibitem[{Fabbri et~al.(2019)Fabbri, Li, She, Li, and Radev}]{fabbri-etal-2019-multi}
Alexander Fabbri, Irene Li, Tianwei She, Suyi Li, and Dragomir Radev. 2019.
\newblock \href {https://doi.org/10.18653/v1/P19-1102} {Multi-news: A large-scale multi-document summarization dataset and abstractive hierarchical model}.
\newblock In \emph{Proceedings of the 57th Annual Meeting of the Association for Computational Linguistics}, pages 1074--1084, Florence, Italy.

\bibitem[{Giorgi et~al.(2022)Giorgi, Soldaini, Wang, Bader, Lo, Wang, and Cohan}]{Giorgi2022TowardsMS}
John Giorgi, Luca Soldaini, Bo~Wang, Gary Bader, Kyle Lo, Lucy~Lu Wang, and Arman Cohan. 2022.
\newblock \href {https://api.semanticscholar.org/CorpusID:258865156} {Towards multi-document summarization in the open-domain}.

\bibitem[{Goyal et~al.(2022)Goyal, Li, and Durrett}]{goyal2022news}
Tanya Goyal, Junyi~Jessy Li, and Greg Durrett. 2022.
\newblock News summarization and evaluation in the era of gpt-3.
\newblock \emph{arXiv preprint arXiv:2209.12356}.

\bibitem[{Hermann et~al.(2015{\natexlab{a}})Hermann, Kociský, Grefenstette, Espeholt, Kay, Suleyman, and Blunsom}]{DBLP:conf/nips/HermannKGEKSB15}
Karl~Moritz Hermann, Tomás Kociský, Edward Grefenstette, Lasse Espeholt, Will Kay, Mustafa Suleyman, and Phil Blunsom. 2015{\natexlab{a}}.
\newblock \href {http://papers.nips.cc/paper/5945-teaching-machines-to-read-and-comprehend} {Teaching machines to read and comprehend}.
\newblock In \emph{NIPS}, pages 1693--1701.

\bibitem[{Hermann et~al.(2015{\natexlab{b}})Hermann, Kočiský, Grefenstette, Espeholt, Kay, Suleyman, and Blunsom}]{hermann-cnndm-2015}
Karl~Moritz Hermann, Tomáš Kočiský, Edward Grefenstette, Lasse Espeholt, Will Kay, Mustafa Suleyman, and Phil Blunsom. 2015{\natexlab{b}}.
\newblock \href {http://arxiv.org/abs/1506.03340} {Teaching machines to read and comprehend}.

\bibitem[{Izacard et~al.(2021)Izacard, Caron, Hosseini, Riedel, Bojanowski, Joulin, and Grave}]{Izacard2021UnsupervisedDI}
Gautier Izacard, Mathilde Caron, Lucas Hosseini, Sebastian Riedel, Piotr Bojanowski, Armand Joulin, and Edouard Grave. 2021.
\newblock \href {https://api.semanticscholar.org/CorpusID:249097975} {Unsupervised dense information retrieval with contrastive learning}.
\newblock \emph{Trans. Mach. Learn. Res.}, 2022.

\bibitem[{Ji et~al.(2013)Ji, Favre, Lin, Gillick, Hakkani-T{\"u}r, and Grishman}]{Ji2013OpenDomainMS}
Heng Ji, Benoit Favre, Wen-Pin Lin, Daniel Gillick, Dilek~Z. Hakkani-T{\"u}r, and Ralph Grishman. 2013.
\newblock \href {https://api.semanticscholar.org/CorpusID:17131989} {Open-domain multi-document summarization via information extraction: Challenges and prospects}.
\newblock In \emph{Multi-source, Multilingual Information Extraction and Summarization}.

\bibitem[{Kulkarni et~al.(2020)Kulkarni, Chammas, Zhu, Sha, and Ie}]{Kulkarni2020AQuaMuSeAG}
Sayali Kulkarni, Sheide Chammas, Wan Zhu, Fei Sha, and Eugene Ie. 2020.
\newblock \href {https://api.semanticscholar.org/CorpusID:225067629} {Aquamuse: Automatically generating datasets for query-based multi-document summarization}.
\newblock \emph{ArXiv}, abs/2010.12694.

\bibitem[{Lewis et~al.(2020{\natexlab{a}})Lewis, Liu, Goyal, Ghazvininejad, Mohamed, Levy, Stoyanov, and Zettlemoyer}]{lewis-etal-2020-bart}
Mike Lewis, Yinhan Liu, Naman Goyal, Marjan Ghazvininejad, Abdelrahman Mohamed, Omer Levy, Veselin Stoyanov, and Luke Zettlemoyer. 2020{\natexlab{a}}.
\newblock \href {https://doi.org/10.18653/v1/2020.acl-main.703} {{BART}: Denoising sequence-to-sequence pre-training for natural language generation, translation, and comprehension}.
\newblock In \emph{Proceedings of the 58th Annual Meeting of the Association for Computational Linguistics}, pages 7871--7880, Online. Association for Computational Linguistics.

\bibitem[{Lewis et~al.(2020{\natexlab{b}})Lewis, Perez, Piktus, Petroni, Karpukhin, Goyal, Kuttler, Lewis, tau Yih, Rockt{\"a}schel, Riedel, and Kiela}]{Lewis2020RetrievalAugmentedGF}
Patrick Lewis, Ethan Perez, Aleksandara Piktus, Fabio Petroni, Vladimir Karpukhin, Naman Goyal, Heinrich Kuttler, Mike Lewis, Wen tau Yih, Tim Rockt{\"a}schel, Sebastian Riedel, and Douwe Kiela. 2020{\natexlab{b}}.
\newblock \href {https://api.semanticscholar.org/CorpusID:218869575} {Retrieval-augmented generation for knowledge-intensive nlp tasks}.
\newblock \emph{ArXiv}, abs/2005.11401.

\bibitem[{Li et~al.(2020)Li, Xiao, Liu, Wu, Wang, and Du}]{Li2020LeveragingGT}
Wei Li, Xinyan Xiao, Jiachen Liu, Hua Wu, Haifeng Wang, and Junping Du. 2020.
\newblock \href {https://api.semanticscholar.org/CorpusID:218718706} {Leveraging graph to improve abstractive multi-document summarization}.
\newblock \emph{ArXiv}, abs/2005.10043.

\bibitem[{Lin(2004)}]{lin-2004-rouge}
Chin-Yew Lin. 2004.
\newblock \href {https://aclanthology.org/W04-1013} {{ROUGE}: A package for automatic evaluation of summaries}.
\newblock In \emph{Text Summarization Branches Out}, pages 74--81, Barcelona, Spain. Association for Computational Linguistics.

\bibitem[{Liu et~al.(2023{\natexlab{a}})Liu, Iter, Xu, Wang, Xu, and Zhu}]{Liu2023GEvalNE}
Yang Liu, Dan Iter, Yichong Xu, Shuo Wang, Ruochen Xu, and Chenguang Zhu. 2023{\natexlab{a}}.
\newblock \href {https://api.semanticscholar.org/CorpusID:257804696} {G-eval: Nlg evaluation using gpt-4 with better human alignment}.
\newblock \emph{ArXiv}, abs/2303.16634.

\bibitem[{Liu and Lapata(2019)}]{Liu2019HierarchicalTF}
Yang Liu and Mirella Lapata. 2019.
\newblock \href {https://api.semanticscholar.org/CorpusID:170079112} {Hierarchical transformers for multi-document summarization}.
\newblock In \emph{Annual Meeting of the Association for Computational Linguistics}.

\bibitem[{Liu et~al.(2023{\natexlab{b}})Liu, Fabbri, Liu, Radev, and Cohan}]{Liu2023OnLT}
Yixin Liu, Alexander~R. Fabbri, Pengfei Liu, Dragomir~R. Radev, and Arman Cohan. 2023{\natexlab{b}}.
\newblock \href {https://api.semanticscholar.org/CorpusID:258841126} {On learning to summarize with large language models as references}.
\newblock \emph{ArXiv}, abs/2305.14239.

\bibitem[{Loshchilov and Hutter(2019)}]{loshchilov2019decoupled}
Ilya Loshchilov and Frank Hutter. 2019.
\newblock \href {http://arxiv.org/abs/1711.05101} {Decoupled weight decay regularization}.

\bibitem[{Lu et~al.(2020)Lu, Dong, and Charlin}]{Lu2020MultiXScienceAL}
Yao Lu, Yue Dong, and Laurent Charlin. 2020.
\newblock \href {https://api.semanticscholar.org/CorpusID:225075639} {Multi-xscience: A large-scale dataset for extreme multi-document summarization of scientific articles}.
\newblock In \emph{Conference on Empirical Methods in Natural Language Processing}.

\bibitem[{Mao et~al.(2023)Mao, Chen, Zhang, Guerin, and Cambria}]{Mao2023GPTEvalAS}
Rui Mao, Guanyi Chen, Xulang Zhang, Frank Guerin, and E.~Cambria. 2023.
\newblock \href {https://api.semanticscholar.org/CorpusID:261100760} {Gpteval: A survey on assessments of chatgpt and gpt-4}.
\newblock \emph{ArXiv}, abs/2308.12488.

\bibitem[{Pasunuru et~al.(2021{\natexlab{a}})Pasunuru, Celikyilmaz, Galley, Xiong, Zhang, Bansal, and Gao}]{Pasunuru2021DataAF}
Ramakanth Pasunuru, Asli Celikyilmaz, Michel Galley, Chenyan Xiong, Yizhe Zhang, Mohit Bansal, and Jianfeng Gao. 2021{\natexlab{a}}.
\newblock \href {https://api.semanticscholar.org/CorpusID:232092834} {Data augmentation for abstractive query-focused multi-document summarization}.
\newblock \emph{ArXiv}, abs/2103.01863.

\bibitem[{Pasunuru et~al.(2021{\natexlab{b}})Pasunuru, Celikyilmaz, Galley, Xiong, Zhang, Bansal, and Gao}]{pasunuru2021data}
Ramakanth Pasunuru, Asli Celikyilmaz, Michel Galley, Chenyan Xiong, Yizhe Zhang, Mohit Bansal, and Jianfeng Gao. 2021{\natexlab{b}}.
\newblock Data augmentation for abstractive query-focused multi-document summarization.
\newblock In \emph{Proceedings of the AAAI Conference on Artificial Intelligence}, volume~35, pages 13666--13674.

\bibitem[{Robertson et~al.(1995)Robertson, Walker, Jones, Hancock-Beaulieu, Gatford et~al.}]{bm25}
Stephen~E Robertson, Steve Walker, Susan Jones, Micheline~M Hancock-Beaulieu, Mike Gatford, et~al. 1995.
\newblock Okapi at trec-3.
\newblock \emph{Nist Special Publication Sp}, 109:109.

\bibitem[{Schick and Sch{\"u}tze(2020)}]{Schick2020ItsNJ}
Timo Schick and Hinrich Sch{\"u}tze. 2020.
\newblock \href {https://api.semanticscholar.org/CorpusID:221703107} {It’s not just size that matters: Small language models are also few-shot learners}.
\newblock \emph{ArXiv}, abs/2009.07118.

\bibitem[{Shen et~al.(2023)Shen, Cheng, You, and Bing}]{Shen2023AreLL}
Chenhui Shen, Liying Cheng, Yang You, and Lidong Bing. 2023.
\newblock \href {https://api.semanticscholar.org/CorpusID:258833685} {Are large language models good evaluators for abstractive summarization?}
\newblock \emph{ArXiv}, abs/2305.13091.

\bibitem[{Tang et~al.(2021)Tang, Sun, Jin, Wang, Zhang, and Wu}]{tang2021improving-pseudo1}
Hongyin Tang, Xingwu Sun, Beihong Jin, Jingang Wang, Fuzheng Zhang, and Wei Wu. 2021.
\newblock \href {http://arxiv.org/abs/2105.03599} {Improving document representations by generating pseudo query embeddings for dense retrieval}.

\bibitem[{Touvron et~al.(2023)Touvron, Martin, Stone, Albert, Almahairi, Babaei, Bashlykov, Batra, Bhargava, Bhosale, Bikel, Blecher, Ferrer, Chen, Cucurull, Esiobu, Fernandes, Fu, Fu, Fuller, Gao, Goswami, Goyal, Hartshorn, Hosseini, Hou, Inan, Kardas, Kerkez, Khabsa, Kloumann, Korenev, Koura, Lachaux, Lavril, Lee, Liskovich, Lu, Mao, Martinet, Mihaylov, Mishra, Molybog, Nie, Poulton, Reizenstein, Rungta, Saladi, Schelten, Silva, Smith, Subramanian, Tan, Tang, Taylor, Williams, Kuan, Xu, Yan, Zarov, Zhang, Fan, Kambadur, Narang, Rodriguez, Stojnic, Edunov, and Scialom}]{Touvron2023Llama2O}
Hugo Touvron, Louis Martin, Kevin~R. Stone, Peter Albert, Amjad Almahairi, Yasmine Babaei, Nikolay Bashlykov, Soumya Batra, Prajjwal Bhargava, Shruti Bhosale, Daniel~M. Bikel, Lukas Blecher, Cristian~Cant{\'o}n Ferrer, Moya Chen, Guillem Cucurull, David Esiobu, Jude Fernandes, Jeremy Fu, Wenyin Fu, Brian Fuller, Cynthia Gao, Vedanuj Goswami, Naman Goyal, Anthony~S. Hartshorn, Saghar Hosseini, Rui Hou, Hakan Inan, Marcin Kardas, Viktor Kerkez, Madian Khabsa, Isabel~M. Kloumann, A.~V. Korenev, Punit~Singh Koura, Marie-Anne Lachaux, Thibaut Lavril, Jenya Lee, Diana Liskovich, Yinghai Lu, Yuning Mao, Xavier Martinet, Todor Mihaylov, Pushkar Mishra, Igor Molybog, Yixin Nie, Andrew Poulton, Jeremy Reizenstein, Rashi Rungta, Kalyan Saladi, Alan Schelten, Ruan Silva, Eric~Michael Smith, R.~Subramanian, Xia Tan, Binh Tang, Ross Taylor, Adina Williams, Jian~Xiang Kuan, Puxin Xu, Zhengxu Yan, Iliyan Zarov, Yuchen Zhang, Angela Fan, Melanie Kambadur, Sharan Narang, Aurelien Rodriguez, Robert Stojnic, Sergey Edunov, and
  Thomas Scialom. 2023.
\newblock \href {https://api.semanticscholar.org/CorpusID:259950998} {Llama 2: Open foundation and fine-tuned chat models}.
\newblock \emph{ArXiv}, abs/2307.09288.

\bibitem[{Wallace et~al.(2020)Wallace, Saha, Soboczenski, and Marshall}]{Wallace2020GeneratingN}
Byron~C. Wallace, Sayantani Saha, Frank Soboczenski, and Iain~James Marshall. 2020.
\newblock \href {https://api.semanticscholar.org/CorpusID:221319573} {Generating (factual?) narrative summaries of rcts: Experiments with neural multi-document summarization}.
\newblock \emph{AMIA ... Annual Symposium proceedings. AMIA Symposium}, 2021:605--614.

\bibitem[{Wang et~al.(2022)Wang, Pang, Chen, Phang, and Bowman}]{wang2022squality}
Alex Wang, Richard~Yuanzhe Pang, Angelica Chen, Jason Phang, and Samuel~R Bowman. 2022.
\newblock Squality: Building a long-document summarization dataset the hard way.
\newblock \emph{arXiv preprint arXiv:2205.11465}.

\bibitem[{Winata et~al.(2021)Winata, Madotto, Lin, Liu, Yosinski, and Fung}]{Winata2021LanguageMA}
Genta~Indra Winata, Andrea Madotto, Zhaojiang Lin, Rosanne Liu, Jason Yosinski, and Pascale Fung. 2021.
\newblock \href {https://api.semanticscholar.org/CorpusID:237532173} {Language models are few-shot multilingual learners}.
\newblock \emph{ArXiv}, abs/2109.07684.

\bibitem[{Xiao et~al.(2021)Xiao, Beltagy, Carenini, and Cohan}]{Xiao2021PRIMERAPM}
Wen Xiao, Iz~Beltagy, Giuseppe Carenini, and Arman Cohan. 2021.
\newblock \href {https://api.semanticscholar.org/CorpusID:247519084} {Primera: Pyramid-based masked sentence pre-training for multi-document summarization}.
\newblock In \emph{Annual Meeting of the Association for Computational Linguistics}.

\bibitem[{Xu and Lapata(2020)}]{Xu2020CoarsetoFineQF}
Yumo Xu and Mirella Lapata. 2020.
\newblock \href {https://api.semanticscholar.org/CorpusID:226262229} {Coarse-to-fine query focused multi-document summarization}.
\newblock In \emph{Conference on Empirical Methods in Natural Language Processing}.

\bibitem[{Yang et~al.(2023)Yang, Li, Zhang, Chen, and Cheng}]{Yang2023ExploringTL}
Xianjun Yang, Yan Li, Xinlu Zhang, Haifeng Chen, and Wei Cheng. 2023.
\newblock \href {https://api.semanticscholar.org/CorpusID:256901227} {Exploring the limits of chatgpt for query or aspect-based text summarization}.
\newblock \emph{ArXiv}, abs/2302.08081.

\bibitem[{Yang et~al.(2015)Yang, Yih, and Meek}]{yang-etal-2015-wikiqa}
Yi~Yang, Wen-tau Yih, and Christopher Meek. 2015.
\newblock \href {https://doi.org/10.18653/v1/D15-1237} {{W}iki{QA}: A challenge dataset for open-domain question answering}.
\newblock In \emph{Proceedings of the 2015 Conference on Empirical Methods in Natural Language Processing}, pages 2013--2018, Lisbon, Portugal. Association for Computational Linguistics.

\bibitem[{Yasunaga et~al.(2017)Yasunaga, Zhang, Meelu, Pareek, Srinivasan, and Radev}]{Yasunaga2017GraphbasedNM}
Michihiro Yasunaga, Rui Zhang, Kshitijh Meelu, Ayush Pareek, Krishna~Parasuram Srinivasan, and Dragomir~R. Radev. 2017.
\newblock \href {https://api.semanticscholar.org/CorpusID:6532096} {Graph-based neural multi-document summarization}.
\newblock \emph{ArXiv}, abs/1706.06681.

\bibitem[{Zhang et~al.(2022{\natexlab{a}})Zhang, Roller, Goyal, Artetxe, Chen, Chen, Dewan, Diab, Li, Lin, Mihaylov, Ott, Shleifer, Shuster, Simig, Koura, Sridhar, Wang, and Zettlemoyer}]{Zhang2022OPTOP}
Susan Zhang, Stephen Roller, Naman Goyal, Mikel Artetxe, Moya Chen, Shuohui Chen, Christopher Dewan, Mona~T. Diab, Xian Li, Xi~Victoria Lin, Todor Mihaylov, Myle Ott, Sam Shleifer, Kurt Shuster, Daniel Simig, Punit~Singh Koura, Anjali Sridhar, Tianlu Wang, and Luke Zettlemoyer. 2022{\natexlab{a}}.
\newblock \href {https://api.semanticscholar.org/CorpusID:248496292} {Opt: Open pre-trained transformer language models}.
\newblock \emph{ArXiv}, abs/2205.01068.

\bibitem[{Zhang et~al.(2020)Zhang, Kishore, Wu, Weinberger, and Artzi}]{BERTScore}
Tianyi Zhang, Varsha Kishore, Felix Wu, Kilian~Q. Weinberger, and Yoav Artzi. 2020.
\newblock \href {https://openreview.net/forum?id=SkeHuCVFDr} {Bertscore: Evaluating text generation with bert}.
\newblock In \emph{International Conference on Learning Representations}.

\bibitem[{Zhang et~al.(2023)Zhang, Ladhak, Durmus, Liang, McKeown, and Hashimoto}]{Zhang2023BenchmarkingLL}
Tianyi Zhang, Faisal Ladhak, Esin Durmus, Percy Liang, Kathleen McKeown, and Tatsunori Hashimoto. 2023.
\newblock \href {https://api.semanticscholar.org/CorpusID:256416014} {Benchmarking large language models for news summarization}.
\newblock \emph{ArXiv}, abs/2301.13848.

\bibitem[{Zhang et~al.(2021)Zhang, Vakulenko, Rajapakse, and Kanoulas}]{Zhang2021ScalingUQ}
Weijia Zhang, Svitlana Vakulenko, Thilina~C. Rajapakse, and E.~Kanoulas. 2021.
\newblock \href {https://api.semanticscholar.org/CorpusID:245131021} {Scaling up query-focused summarization to meet open-domain question answering}.
\newblock \emph{ArXiv}, abs/2112.07536.

\bibitem[{Zhang et~al.(2022{\natexlab{b}})Zhang, Ni, Mao, Wu, Zhu, Deb, Awadallah, Radev, and Zhang}]{zhang-etal-2022-summn}
Yusen Zhang, Ansong Ni, Ziming Mao, Chen~Henry Wu, Chenguang Zhu, Budhaditya Deb, Ahmed Awadallah, Dragomir Radev, and Rui Zhang. 2022{\natexlab{b}}.
\newblock \href {https://doi.org/10.18653/v1/2022.acl-long.112} {{S}umm$^n$: A multi-stage summarization framework for long input dialogues and documents}.
\newblock In \emph{Proceedings of the 60th Annual Meeting of the Association for Computational Linguistics (Volume 1: Long Papers)}, pages 1592--1604, Dublin, Ireland. Association for Computational Linguistics.

\bibitem[{Zhao et~al.(2023)Zhao, Qi, Nan, Mi, Liu, Zou, Han, Tang, Xu, Cohan et~al.}]{zhao2023qtsumm}
Yilun Zhao, Zhenting Qi, Linyong Nan, Boyu Mi, Yixin Liu, Weijin Zou, Simeng Han, Xiangru Tang, Yumo Xu, Arman Cohan, et~al. 2023.
\newblock \href {https://arxiv.org/abs/2305.14303} {Qtsumm: A new benchmark for query-focused table summarization}.
\newblock \emph{arXiv preprint arXiv:2305.14303}.

\bibitem[{Zhong et~al.(2021)Zhong, Yin, Yu, Zaidi, Mutuma, Jha, Awadallah, Celikyilmaz, Liu, Qiu, and Radev}]{Zhong2021QMSumAN}
Ming Zhong, Da~Yin, Tao Yu, Ahmad~Zairi Zaidi, Mutethia Mutuma, Rahul Jha, Ahmed~Hassan Awadallah, Asli Celikyilmaz, Yang Liu, Xipeng Qiu, and Dragomir~R. Radev. 2021.
\newblock \href {https://api.semanticscholar.org/CorpusID:233219904} {Qmsum: A new benchmark for query-based multi-domain meeting summarization}.
\newblock In \emph{North American Chapter of the Association for Computational Linguistics}.

\end{thebibliography}

\newpage

\appendix

\section*{Appendix}
\label{sec:appendix}

\section{q2OD-MDS}
\label{sec:q2ODMDS}
Here is a mathematical definition for q2OD-MDS:
\begin{equation}
\fbox{%
  \begin{minipage}{0.3\textwidth}
  \begin{align*}
  &\textbf{qMDS:} \quad \{(q_i, D_i, s_i)\}_{i=1}^n \\
  &\textbf{where:} \\
  &q_i \text{ is the } i\text{-th query,} \\
  &D_i \text{ is the set of documents related to } q_i, \\
  &s_i \text{ is the summary related to } q_i. \\
  &\textbf{Transformation:} \\
  &\textbf{Step 1: Extract query-summary pairs:}\\
  &\quad \{(q_i, s_i)\}_{i=1}^n \\
  &\textbf{Step 2: Combine all documents:} \quad D = \bigcup_{i=1}^n D_i \\
  &\textbf{ODMDS:} \quad (D, \{(q_i, s_i)\}_{i=1}^n)
  \end{align*}
  \end{minipage}
}
\end{equation}

\begin{table}[!t]
\small
\begin{center}
\begin{tabular}{ lrrr } 
 \toprule
 Method & Top-\textit{k} Strategy & P@K & R@K \\
 \midrule
 Original & min(3) & 0.48 & 0.25 \\
& mean(8) & 0.36  & 0.34 \\
 & max(10) & 0.32  & 0.33 \\
  With title & min & 0.66 & 0.35 \\
& mean & 0.49  & 0.46 \\
 & max & 0.42  & 0.45 \\
  With title and answer & min & \textbf{0.74} & 0.40 \\
& mean & 0.54  & \textbf{0.52} \\
 & max & 0.47  & 0.50 \\
\bottomrule
\end{tabular}
\end{center}
\caption{Performances using BM25 with different modifications of queries.}

\label{tab:query-mod}
\end{table}

\section{Transforming QMSum}
\label{math-def-of-qmsum}
We formalize the process of transforming QMSum into a new dataset mathematically as follows:
\begin{itemize}
\item \textbf{Meetings Transcripts:} Denote the set of all meetings transcripts as $T = \{T_1, T_2, \ldots, T_n\}$, where $T_i$ represents the transcript of the $i$-th meeting.

\item \textbf{Queries \& Summaries:} Each meeting transcript $T_i$ has an associated set of queries $Q_i = \{q_{i1}, q_{i2}, \ldots, q_{im}\}$, where $q_{ij}$ represents the $j$-th query in the $i$-th meeting. Each query $q_{ij}$ has a corresponding summary $s_{ij}$. Therefore, we have a set of summaries $S_i = \{s_{i1}, s_{i2}, \ldots, s_{im}\}$ for each meeting transcript $T_i$.

\item \textbf{Query Clustering:} Denote the function that measures the similarity between two queries as $\text{sim}(q_{ij}, q_{kl})$, where $q_{ij}$ and $q_{kl}$ are queries from meetings $i$ and $k$, respectively. Here, we used \textit{text-embedding-ada-002} for the purpose of implementing $\text{sim}(q_{ij}, q_{kl})$. We then define a threshold $\theta$ such that if $\text{sim}(q_{ij}, q_{kl}) > \theta$, we consider $q_{ij}$ and $q_{kl}$ to be in the same cluster. The result of this step is a new set of query clusters $C = \{C_1, C_2, \ldots, C_p\}$, where each cluster $C_k = \{q_{ij} | \text{sim}(q_{ij}, q_{kl}) > \theta \text{ for all } q_{kl} \text{ in } C_k\}$. 

\item \textbf{Cluster Modification:} We ensured that each cluster has a minimum and maximum size, which can be useful for ensuring that each cluster is meaningful (not too small) and manageable (not too large). Here we choose $max$ to be 6 and $min$ to be 2. For any cluster $C_k$ with more than $max$ queries, we performed a secondary clustering operation within $C_k$ using a lower similarity threshold. Similarly, for any cluster $C_k$ with less than $min$ queries, we merge it with the most similar other cluster. After this step, we have a modified set of clusters $C' = \{C'_1, C'_2, \ldots, C'_q\}$.

\item \textbf{Merging:} For each modified cluster $C'_k$, a new merged query $Q'_k$ and a corresponding summary $S'_k$ are created by combining the queries and summaries in the cluster, respectively. Here we also asked LLM to merge them together for us. After this step, we have a set of merged query-summary pairs $\{(Q'_1, S'_1), (Q'_2, S'_2), \ldots, (Q'_q, S'_q)\}$, where each pair corresponds to a modified cluster.

\end{itemize}

The final result of this process is a new set of query-summary pairs $\{(Q_k, S_k)\}$, providing a condensed overview of multiple meetings.

\section{Prompt Templates}
\label{sec:template}

\subsection{Summarization Prompt Templates}
\begin{itemize}

\item \textbf{ODSum-story}:\\\textit{You are a helpful assistant that gives long answer to question based on a long story. Write an answer based on the following question and the story. STORY:\{story\} QUESTION:\{query\} SUMMARY:}\\ Here \textit{STORY} represents a selection of the most matching stories retrieved by different retrieval models based on the query from ODSum-story.

\item \textbf{ODSum-meeting}: \\\textit{You are a helpful assistant that gives long answer to question based on a long meeting. Write an answer based on the following question and the given meeting. Try to answer thoroughly and do not leave out useful information. MEETING:\{meeting\} QUESTION:\{query\} SUMMARY:} \\In this case, \textit{MEETING} refers to a selection of the most matching stories retrieved by different retrieval models based on the query from ODSum-meeting.

\subsection{G-E\small{VAL} Prompt Templates}
\item \textbf{G-E{\small VAL} consistence:} \\
\textit{You will be given a news article. You will then be given one summary written for this article. Your task is to rate the summary on one metric. Please make sure you read and understand these instructions carefully. Please keep this document open while reviewing, and refer to it as needed. Evaluation Criteria: Consistency (1-5) - the factual alignment between the summary and the summarized source. A factually consistent summary contains only statements that are entailed by the source document. Annotators were also asked to penalize summaries that contained hallucinated facts. }

\textit{Evaluation Steps: 1. Read the news article carefully and identify the main facts and details it presents. 2. Read the summary and compare it to the article. Check if the summary contains any factual errors that are not supported by the article. 3. Assign a score for consistency based on the Evaluation Criteria.}
\item \textbf{G-E{\small VAL} relevance:} \\
\textit{You will be given one summary written for a news article. Your task is to rate the summary on one metric. Please make sure you read and understand these instructions carefully. Please keep this document open while reviewing, and refer to it as needed. Evaluation Criteria: Relevance (1-5) - selection of important content from the source. The summary should include only important information from the source document. Annotators were instructed to penalize summaries which contained redundancies and excess information. }

\textit{Evaluation Steps: 1. Read the summary and the source document carefully. 2. Compare the summary to the source document and identify the main points of the article. 3. Assess how well the summary covers the main points of the article, and how much irrelevant or redundant information it contains. 4. Assign a relevance score from 1 to 5.}

\end{itemize}

\section{Dataset Example}
\label{sec:dataset_Example}

We show one example of each setting of the ODSum dataset in Table \ref{tab:squality_example} and Table \ref{tab:qmsum_example}.

\begin{table*}[t!]
\footnotesize
\fontfamily{ppl}\selectfont
\begin{tabular}{l|p{12cm}}  %
 \toprule
 \textbf{\textit{Field}} & Content \\
 \midrule
  \textbf{\textit{Query}} & What is the role of \textcolor{red}{Hank Arapoulous} in the story Cultural Exchange? \\
 \hline
  \textbf{\textit{Retrieved\_document\_1}} & ...What can I do for you? Retief said.     You work for this Culture bunch, do you? Funny. I thought they wereall ribbon-counter boys. Never mind. I'm \textcolor{red}{Hank Arapoulous}. \textcolor{blue}{I'm a farmer}.What I wanted to see you about was... He shifted in his chair. Well, \textcolor{blue}{out on Lovenbroy we've got a serious problem}. The wine crop is justabout ready. We start picking in another two, three months. Now I don'tknow if you're familiar with the Bacchus vines we grow...?     No, Retief said. Have a cigar? He pushed a box across the desk. \textcolor{red}{Arapoulous} took one. Bacchus vines are an unusual crop, he said,puffing the cigar alight. Only mature every twelve years. In between,the vines don't need a lot of attention, so our time's mostly our own.We like to farm, though. Spend a lot of time developing new forms. Apples the size of a melon... \\
 \hline
  \textbf{\textit{Retrieved\_document\_2}} & Retief lay on his back in deep grass by a stream, eating grapes. A tallfigure appeared on the knoll above him and waved.     Retief! \textcolor{red}{Hank Arapoulous} bounded down the slope and embraced Retief,slapping him on the back. I heard you were here, and I've got news for you. \textcolor{blue}{You won the final day's picking competition}. Over two hundredbushels! That's a record!     Let's get on over to the garden. Sounds like the celebration's aboutto start.     In the flower-crowded park among the stripped vines, Retief and \textcolor{red}{Arapoulous} made their way to a laden table under the lanterns. A tallgirl dressed in loose white, and with long golden hair, came up to \textcolor{red}{Arapoulous}.     Delinda, this is Retief today's winner. And he's also the fellow that got those workers for us... \\
 \hline
 \textbf{\textit{Retrieved\_document\_3}} & The secretary placed the papers on the desk. \textcolor{red}{Arapoulous} caught her eyeand grinned. She sniffed and marched from the room.     What that gal needs is a slippery time in the grape mash, \textcolor{red}{Arapoulous} observed. Retief thumbed through the papers, pausing to read from time to time. He finished and looked at \textcolor{red}{Arapoulous}.     How many men do you need for the harvest, Hank? Retief inquired.     \textcolor{red}{Arapoulous} sniffed his wine glass and looked thoughtful.     A hundred would help, he said. A thousand would be better. Cheers.   \textcolor{blue}{  What would you say to two thousand?}     Two thousand? Retief, you're not fooling?     I hope not. He picked up the phone, called the Port Authority, asked for the dispatch clerk.     Hello, Jim. Say, I have a favor to ask of you. You know that contingent of Bogan students. They're traveling aboard the two CDT transports. I'm interested in the baggage that goes with the students.Has it arrived yet? Okay, I'll wait... \\
 \hline
\textbf{\textit{Summary\_1}}&   \textcolor{red}{Hank Arapoulous} is first described as \textcolor{blue}{a bucolic person from Lovenbroy}.\textcolor{blue}{He is a farmer, tall with bronze skin and gray hair}, who comes to MUDDLE's office \textcolor{blue}{to discuss the harvest problems in Lovenbroy}. They grow Bacchus vines, which only mature once every twelve years. This year is a harvest year, but they don't have enough people to harvest the grapes. Arapoulous explains to Retief that a few years ago, Boge landed a force on Lovenbroy to try to mine their minerals by strip-mining. Lovenbroy fought back for a year but lost a lot of its men. This created financial problems, so Lovenbroy borrowed money from Croanie, mortgaging its crops. The loan is due, and the wine crop will cover the loan amount, but they don't have enough people to harvest the grapes. He is worried that if they don't have a great harvest, Croanie will come in and start mining. Also, if they default on the loan, Croanie will hold half of the grape acreage that they used to secure the loan. Arapoulous has also asked for help from the Labor Office, but they only offered to send them machinery, and machines cannot harvest the grapes. He returns to see Retief the following day to find out if Retief has discovered a way to help. When Mr. Karsh makes a scene about the missing luggage for the exchange students, Retief has Arapoulous take Karsh away and take care of him. When they return, Karsh is stumbling and needs support to stand up. Arapoulous explains that Karsh fell. \textcolor{blue}{Retief sends the exchange students to Lovenbroy with Arapoulous to help with the harvest}. As the harvest is winding down, \textcolor{blue}{Arapoulous tells Retief that Retief has won the award for the picking competition}. Arapoulous is also the person who judges the wine contest. \\
\hline
\textbf{\textit{Summary\_2}}& ...\\

\bottomrule
\end{tabular}

\caption{Example from the ODSum-story Dataset}
\label{tab:squality_example}
\end{table*}

\begin{table*}[t!]
\footnotesize
\fontfamily{ppl}\selectfont
\begin{tabular}{l|p{12cm}}  %
 \toprule
 \textbf{\textit{Field}} & Content \\
 \midrule
  \textbf{\textit{Query}} & What were the changes needed according to the group and the Professor's opinion on \textcolor{red}{the use of English} in the discussion about IBM computers and \textcolor{red}{data collection participants}? \\
 \hline
  \textbf{\textit{Retrieved\_document\_1}} & 
  ...People always say very glibly that if you show improvement on a bad system , that doesn't mean anything , cuz it may not be show  , because , you know , it doesn't tell you anything about the good system . You know , that if some people If you 're actually are getting at something that has some conceptual substance to it , it will port . professor b: And in fact , most methods that people now use were originally tried with something that was not their absolute best system at some level . \textcolor{blue}{If we 're getting three percent error on, English , native speakers , using the Aurora system , and we do some improvements and bring it from three to two , do those same improvements bring, you know, the SRI system from one point three to you know, to point eight} ? professor b: You know , that 's that 's pretty solid , on the segmentation stuff . And the Aurora folks here will will definitely get something in on Aurora , phd d: which is not phd f: Actually this this,  So, there 's another paper. phd f: And he tested it mostly on digits because it 's sort of a you know , it doesn't take weeks to train it. And got some very impressive results, with, you know , discriminative , Gaussian training. you know, like, error rates go from I don't know , in very noisy environment , like from , ...
 \\
 \hline
  \textbf{\textit{Retrieved\_document\_2}} & ...without the adaptation and compare to these numbers without the adaptation. but I 'm not so much worried about the adaptation , actually , than the , VTL estimation . phd f: If you have only one utterance per speaker you might actually screw up on estimating the the warping , factor . But it 's not the amount of speakers , it 's the num it 's the amount of data per speaker. grad e: So , although I sort of know how to run it , there are a little a f few details here and there that I 'll have to dig out . phd f: And there 's a script and that is actually all in one script . So there 's this one script that parses waveform names and extracts things like the , speaker , ID or something that can stand in as a speaker ID . \textcolor{blue}{So , we might have to modify that script to recognize the , speakers , in the in the , TI - digits database}. phd f: Or you can fake you can fake names for these waveforms that resemble the names that we use here for the for the meetings . phd f: That would be the , sort of probably the safest way to do grad e: I might have to do that anyway to to do because we may have to do an extract to get the amount of data per speaker about right . grad e: The other thing is , isn't TI - digits isolated digits ? phd f: Right . grad e: Or is that another one ? I looked through a bunch of the digits t corp corpora , and now they 're all blurring . phd f: By the way , I think we can improve these numbers if we care to compr improve them by , not starting with the Switchboard models but by taking the Switchboard models and doing supervised adaptation on a small amount of digit data collected in this setting . \\

 \hline
\textbf{\textit{Summary}}&   The script for extracting speaker ID information needs to be modified in order to improve the accuracy of data estimation per speaker. The professor believes that \textcolor{blue}{removing English did not significantly impact the model}, and it would be valuable to train on one language and test on another. The team expressed concerns about the availability of computational resources, but the professor informed them that they had received two 550 megahertz processors from IBM instead of the expected eight 800 megahertz processors. Currently, there are already 30 subjects lined up for the study, with more expected to be recruited off campus. It was agreed that having subjects select tasks from different categories, such as "big place" or "service", would provide a better range of data. Each dialogue is expected to last no more than 10 minutes. As the data collection is about to begin, \textcolor{blue}{minor changes need to be made to the experiment design, script, and permission forms}. Subjects can be recruited from within the university or through other social circles. \\
\hline
\textbf{\textit{BART\_generated\_Summary}} & The group discussed the possibility of using IBM computers for data collection, with a focus on the performance of the H T K system, which was used by all participants in Aurora, Colorado. They also discussed the use of English as a language in the data collection process, with \textcolor{blue}{the professor expressing concerns about the potential loss of English due to lack of usage}. However, the professor emphasized the importance of incorporating English into data collection and emphasized the need for the team to learn English properly... \\
\hline
\textbf{\textit{gpt-3.5\_generated\_Summary}} &  \\

\bottomrule
\end{tabular}

\caption{Example from the ODSum-meeting Dataset}
\label{tab:qmsum_example}
\end{table*}

\end{document}